\documentclass{article}

\usepackage{microtype}
\usepackage{graphicx}
\usepackage{booktabs} 
\usepackage{subcaption}
\usepackage{hyperref}

\usepackage[accepted]{icml2021}

\icmltitlerunning{DP-SGD vs PATE: Which Has Less Disparate Impact on Model Accuracy?}

\begin{document}

\twocolumn[
\icmltitle{DP-SGD vs PATE: Which Has Less Disparate Impact on Model Accuracy?}



\icmlsetsymbol{equal}{*}

\begin{icmlauthorlist}
\icmlauthor{Archit Uniyal}{equal,panj,om}
\icmlauthor{Rakshit Naidu}{equal,om,man,cmu}
\icmlauthor{Sasikanth Kotti}{iitj,om} \\
\icmlauthor{Sahib Singh}{om,ford} 
\icmlauthor{Patrik Joslin Kenfack}{inno} \\
\icmlauthor{Fatemehsadat Mireshghallah}{ucsd,om} 
\icmlauthor{Andrew Trask}{ou,om}
\end{icmlauthorlist}

\icmlaffiliation{panj}{Panjab University}
\icmlaffiliation{man}{Manipal Institute of Technology}
\icmlaffiliation{cmu}{Carnegie Mellon University}
\icmlaffiliation{iitj}{IIT Jodhpur}
\icmlaffiliation{inno}{Innopolis University}
\icmlaffiliation{ucsd}{University of California, San Diego}
\icmlaffiliation{om}{OpenMined}
\icmlaffiliation{ou}{University of Oxford}
\icmlaffiliation{ford}{Ford Motor Company}

\icmlcorrespondingauthor{Sahib Singh}{sahibsingh570@gmail.com}
\icmlcorrespondingauthor{Fatemehsadat Mireshghallah}{fmireshg@eng.ucsd.edu}

\icmlkeywords{Machine Learning, ICML}

\vskip 0.3in
]



\printAffiliationsAndNotice{\icmlEqualContribution} 

\begin{abstract}
Recent advances in differentially private deep learning have demonstrated that application of differential privacy-- specifically the DP-SGD algorithm-- has a disparate impact on different sub-groups in the population, which leads to a significantly high drop-in model utility for sub-populations that are under-represented (minorities), compared to well-represented ones. In this work, we aim to compare PATE, another mechanism for training deep learning models using differential privacy, with DP-SGD in terms of fairness. We show that PATE does have a disparate impact too, however, it is much less severe than DP-SGD. We draw insights from this observation on what might be promising directions in achieving better fairness-privacy trade-offs.  

\end{abstract}

\section{Introduction}

There are a plethora of attacks that exploit the vulnerabilities in  machine learning models and infer membership of certain individuals~\cite{memberinf, jayaraman2020revisiting} or   infer sensitive attributes such as age, ethnicity or genetic markers of given data instances, from released models in both black and white-box settings ~\cite{shokri, pharmacogenetics}. 
One approach to protect against such attacks is Differential Privacy (DP). DP can protect against attempts to infer the contribution of a given individual to the training set by adding noise to the computations~\cite{chaudhuriERM, abadi2016deep, Mireshghallah2020PrivacyID, jayaraman2019evaluating}.

A widely adopted DP mechanism in deep learning is DP-Stochastic Gradient Descent (DP-SGD)~\cite{abadi2016deep} wherein noise is added to clipped gradients during training. 
Although DP-SGD offers rigorous privacy guarantees, it degrades the accuracy of the resulting model. 
Furthermore, it has recently been shown that this degradation is disparate, and that differential privacy exacerbates the already existing gap between the utility of the model for under-represented and well-represented subgroups of data~\cite{bagdasaryan2019differential, kuppam2019fair, farrand2020neither}. 
Given how machine learning  models are deployed in real life and are sometimes used to make high-stake decisions such as who to hire for a role or how much someone should pay for insurance, it is paramount that we find, acknowledge and mitigate biases induced by different algorithms, in this case DP. 

To this end, we aim at studying the fairness implications of yet 
another DP mechanism for training deep neural networks, named PATE~\cite{papernot2016semi}. In PATE (Private Aggregation of Teacher Ensembles), sensitive data is split into a certain number of disjoint training sets, on each of which a teacher classifier is trained. Then, a student model is trained by transferring the  noisy aggregate knowledge of the teachers. The noisy aggregation is where the privacy guarantees are induced through DP.
%


While DP-SGD has been studied with regards to its disparate impact on imbalanced data, there has been no rigorous study on how PATE performs in such scenarios. In this work, we seek to extensively compare the utility loss across DP-SGD and PATE on imbalanced datasets, where there are under-represented sub-populations. Through outlining the performance of both these approaches across varying levels of privacy.
We seek to answer the following question: 
``Do both DP-SGD and PATE have similar disparate impacts on minority classes?''
To answer briefly, through our experiments with MNIST and SVHN datasets, we observe that PATE has significantly higher utility on under-represented groups (i.e. lower accuracy parity), which we attribute to its teacher-ensemble setup. The code for this paper can be found at~\href{https://github.com/sahibsin/Private_ClassImbalance/}{DP-SGD vs PATE Github repository}.

\section{Related Work}

It has been shown in the literature that training models with the ultimate goal of maximizing accuracy leads to learning and even amplifying the biases in data. 
Bias in machine learning can be defined as the phenomena of observing results which are systematically prejudiced due to faulty assumptions.
The survey by~\citeauthor{mehrabi2019survey} covers various types of bias, such as historic bias, representation bias and algorithmic bias. 
Differential privacy is used in many contexts, including but not limited to healthcare settings~\cite{singh2020benchmarking, suriyakumar2021chasing}, commerce (US Census data release)~\cite{abowd2018us, fioretto2021differential} and natural language processing applications such as next word prediction in keyboards~\cite{kairouz2021practical}. 

The works by~\citeauthor{bagdasaryan2019differential} and ~\citeauthor{farrand2020neither} empirically demonstrates that when DP-SGD is used on highly imbalanced class data the less represented groups which already have lower accuracy end up losing more utility: ``the poor become poorer". They also show that as stricter privacy guarantees are imposed, this gap gets wider.
Our work is similar in nature to these works, however we seek to study another differentially private learning mechanism, named PATE (Private aggregation of Teacher Ensembles, ~\citet{papernot2016semi}) and compare it to DP-SGD in terms of the parity in accuracy of different subgroups.
~\citeauthor{kuppam2019fair} and ~\citeauthor{Fioretto2021DecisionMW} show that if DP was used in the decision making of a fund allocation problem (based on US Census data), smaller districts would end up with comparatively more funding than what they would receive without DP, and larger districts would get less funding.
~\citeauthor{jagielski2018differentially} propose differentially private variants of existing bias mitigation techniques for creating fair classifiers.~\citeauthor{tran2020differentially} propose a differentially private learning algorihtm, with fairness constraints imposed.

\section{Differential Privacy for Deep Learning}

Here we discuss the main privacy concepts used in the paper.

\subsection{Differentially Private SGD (DP-SGD)}
For  $\epsilon  \geq 0$, an algorithm $A$ is understood to be Differentially Private \cite{dwork2006our,dwork2006calibrating} if and only if for any pair of datasets that differ in only one element, the following statement holds true.
\begin{center}
$P[A(D) = t] \leq e^\epsilon$ $P[A(D^{'}) = t] $ $\; \forall t$
\end{center}
Where $D$ and $D'$ are differing datasets by at most one element, and $P[A(D) = t]$ denotes the probability that $t$ is the output of $A$. This setting approximates the effect of individual opt-outs by minimising the inclusion effect of an individual’s data. 

DP-SGD \cite{abadi2016deep} is a modification of the stochastic gradient descent algorithm which provides provable privacy guarantees. DP-SGD bounds the sensitivity of each gradient and is paired with a moments accountant algorithm to amplify and track the privacy loss across weight updates. Moments accountant significantly improves on earlier privacy analysis of SGD and allows for meaningful privacy guarantees for deep learning trained on realistically sized datasets \cite{bu2019deep}.

\subsection {Private Aggregation of Teacher Ensembles (PATE)}
PATE  \cite{papernot2016semi} is based on knowledge aggregation and transfer from “teacher” models, trained on disjoint data, to a “student” model whose attributes may be made public. 
The approach involves transferring knowledge of  multiple models trained on sensitive disjoint datasets. Since these models rely directly on sensitive data, they are not published, but are used as “teachers” for a “student” model. The student learns to predict an output chosen by noisy voting among all of the teachers, and cannot directly access an individual teacher or the underlying data or parameters.
The student is trained on a publicly available, unlabeled dataset, where the labels come from the aggregate votes of the teachers. That is why PATE is considered a semi-supervised approach, and it cannot be applied as it is in as setup where we do not have public data (even small in number) available.


\begin{figure*}[h]
    \centering
        \begin{subfigure}{0.25\textwidth}
     \includegraphics[width=\linewidth]{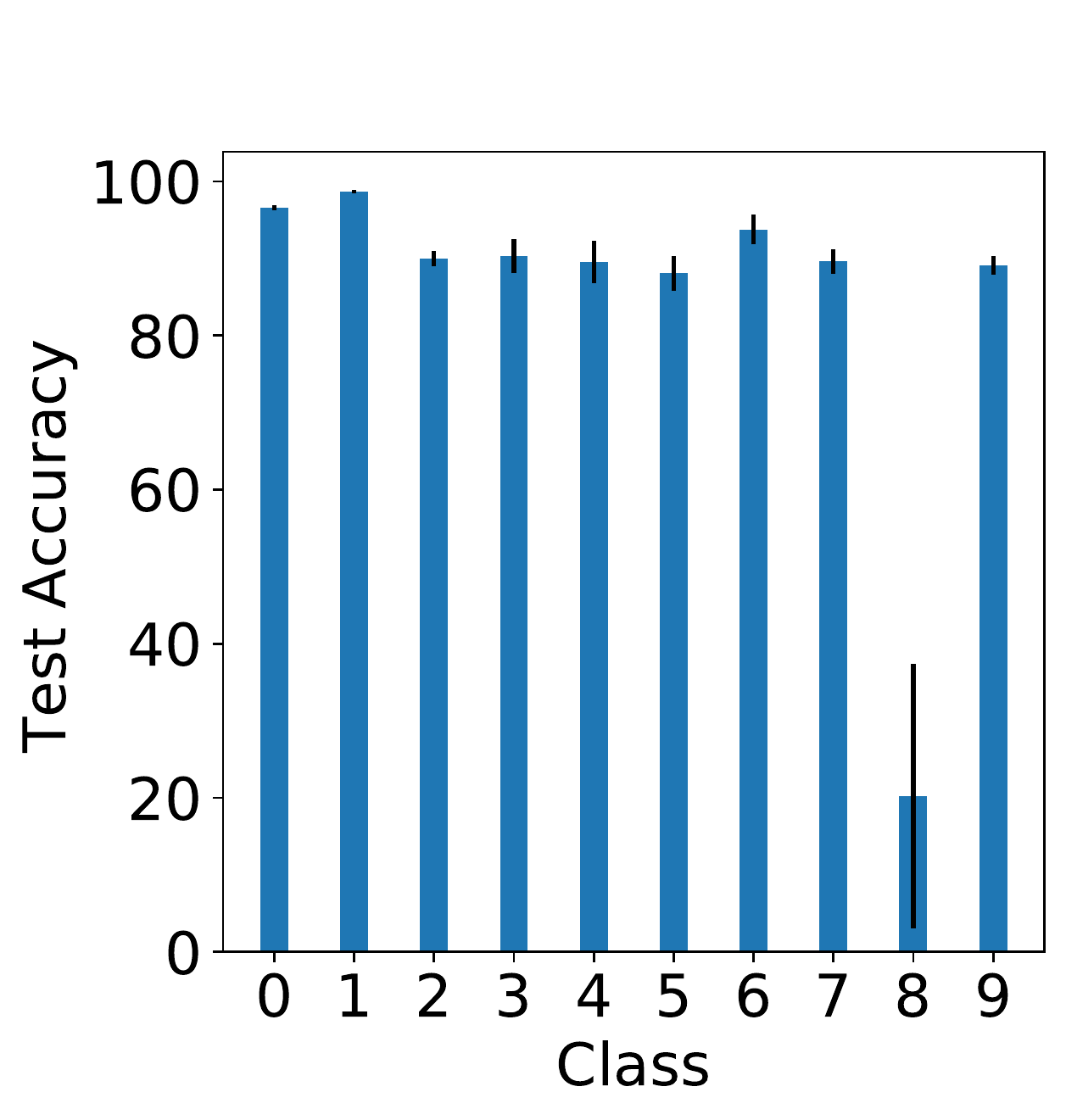}
     \caption{DP-SGD for $\varepsilon=0.5$}
     \label{fig:mnistdpsgdeps0.5}
    \end{subfigure}
    \begin{subfigure}{0.25\textwidth}
     \includegraphics[width=\linewidth]{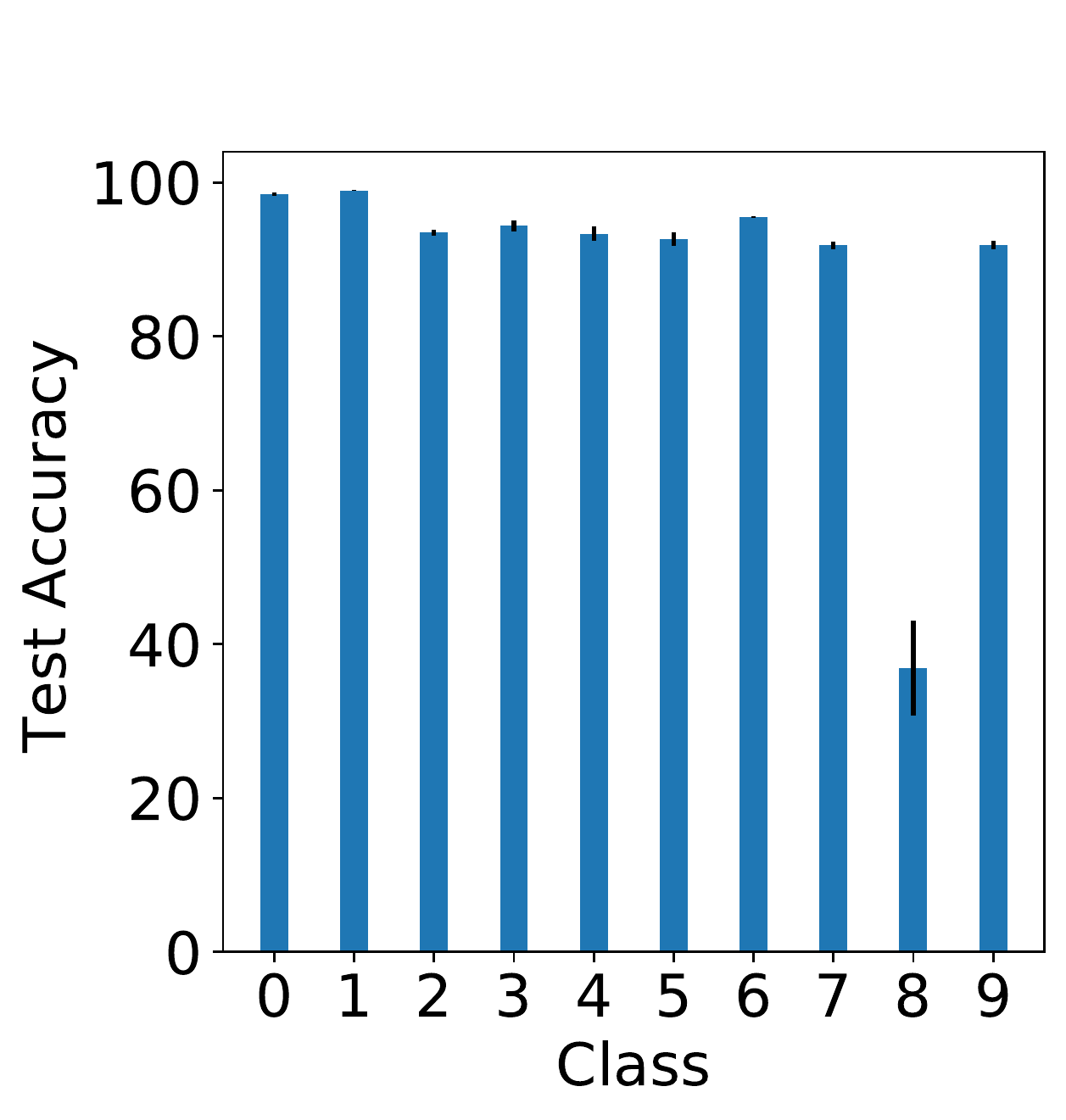}
     \caption{DP-SGD for $\varepsilon=5$}
     \label{fig:mnistdpsgdeps5}
    \end{subfigure}
    \begin{subfigure}{0.25\textwidth}
     \includegraphics[width=\linewidth]{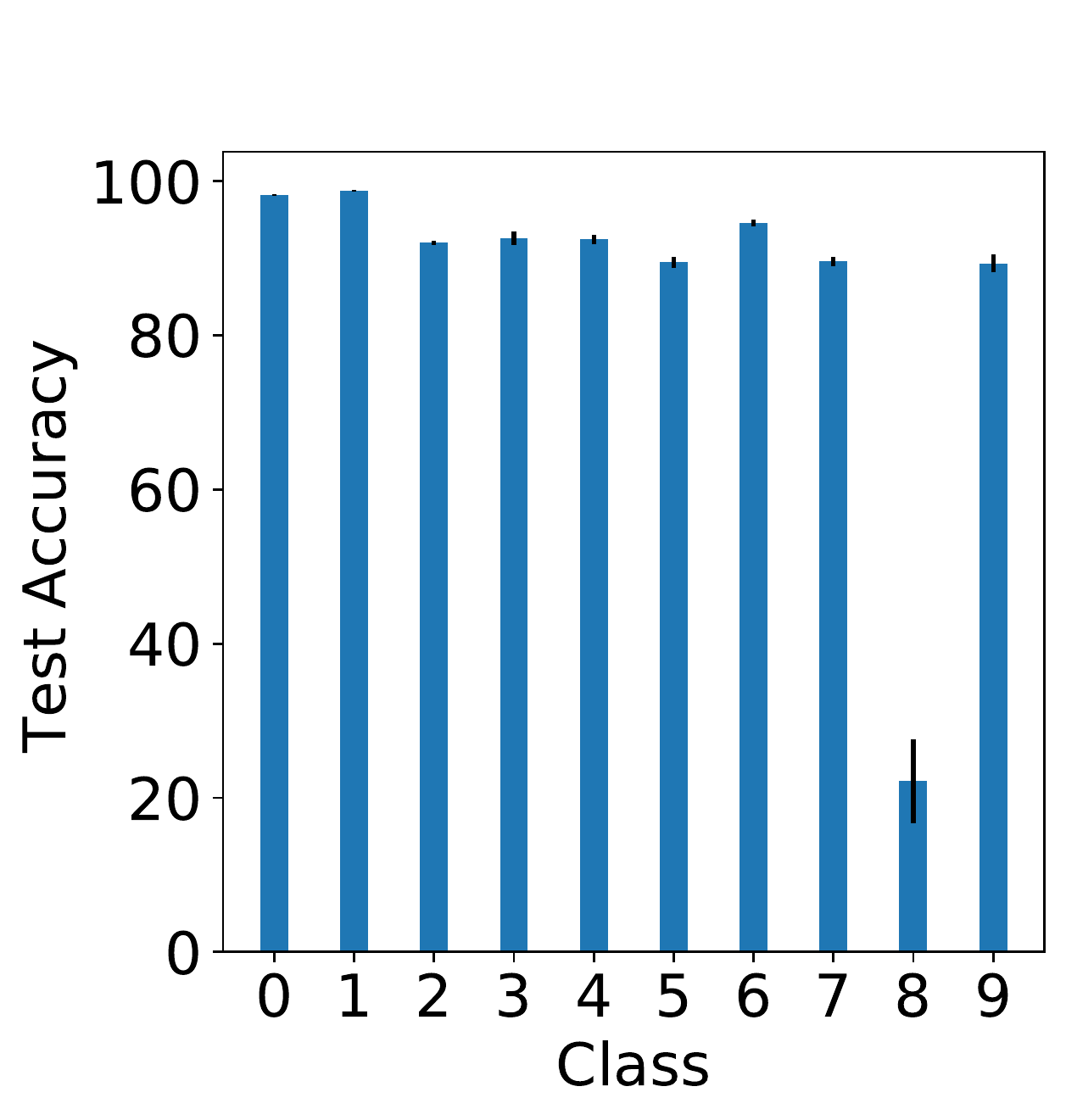}
     \caption{DP-SGD for $\varepsilon=15$}
     \label{fig:mnistdpsgdeps15}
    \end{subfigure}
    
        \begin{subfigure}{0.25\textwidth}
     \includegraphics[width=\linewidth]{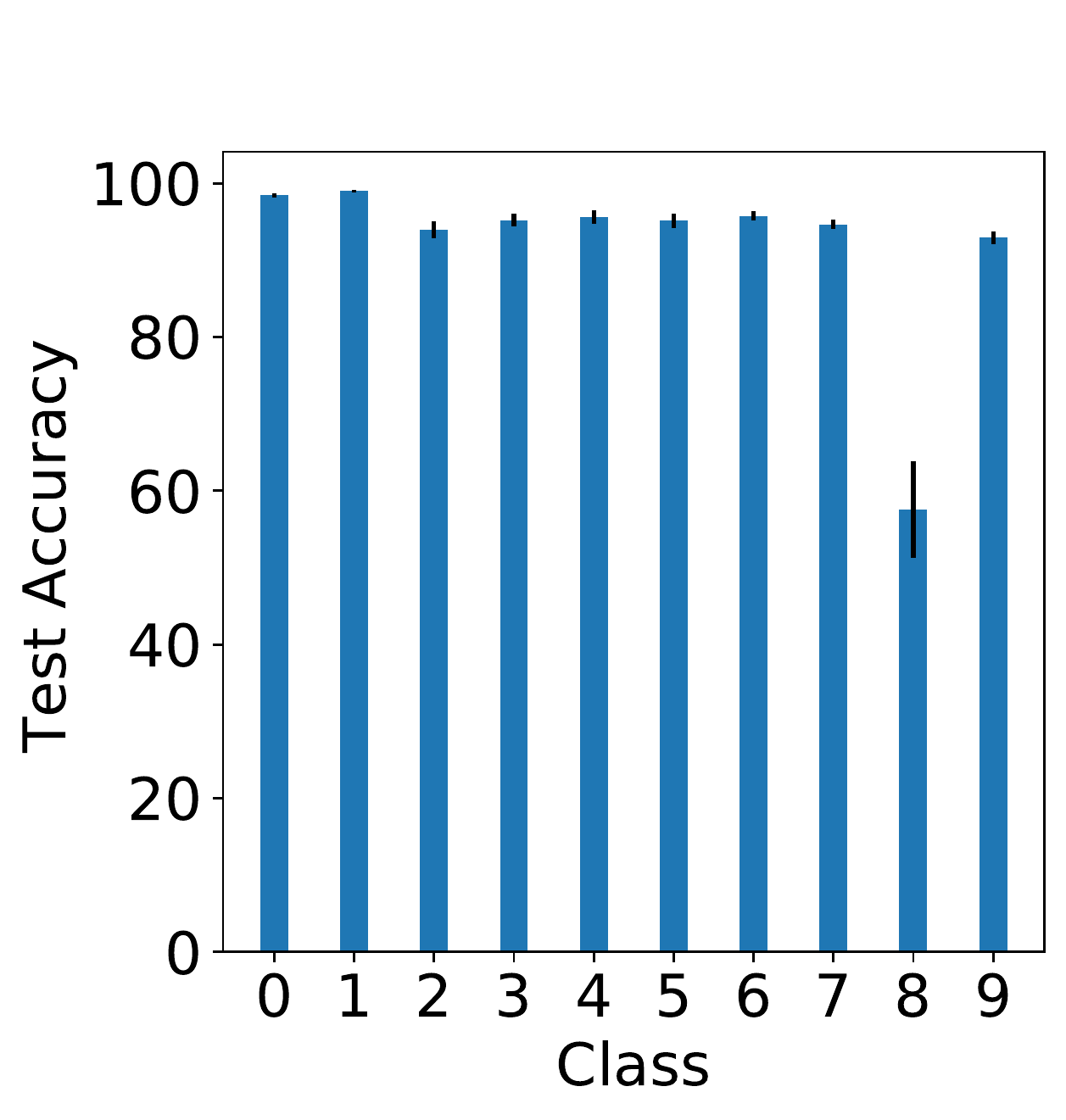}
     \caption{PATE for $\varepsilon=0.5$}
     \label{fig:mnistpateeps0.5}
    \end{subfigure}
    \begin{subfigure}{0.25\textwidth}
     \includegraphics[width=\linewidth]{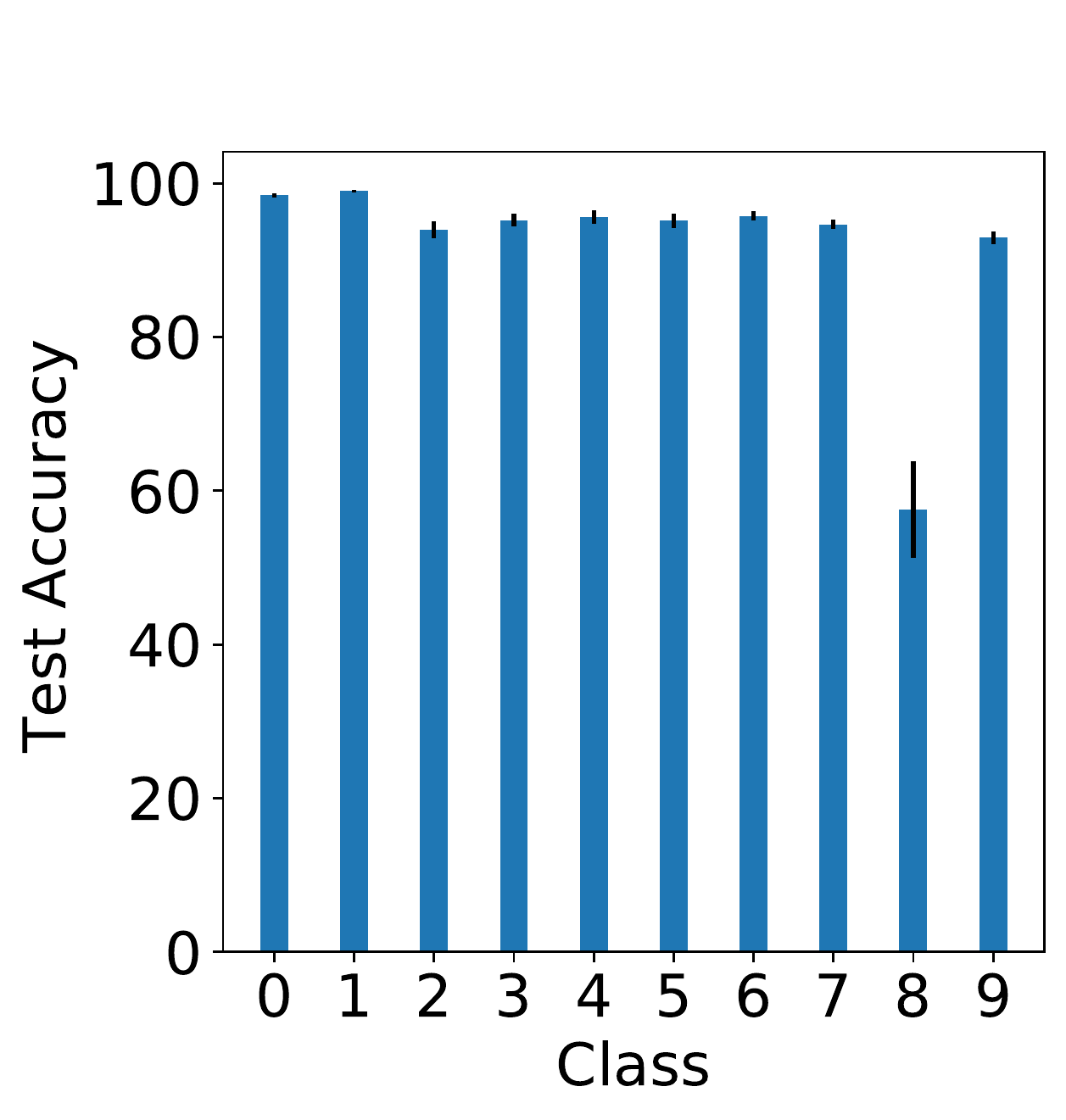}
     \caption{PATE for $\varepsilon=5$}
     \label{fig:mnistpateeps5}
    \end{subfigure}
    \begin{subfigure}{0.25\textwidth}
     \includegraphics[width=\linewidth]{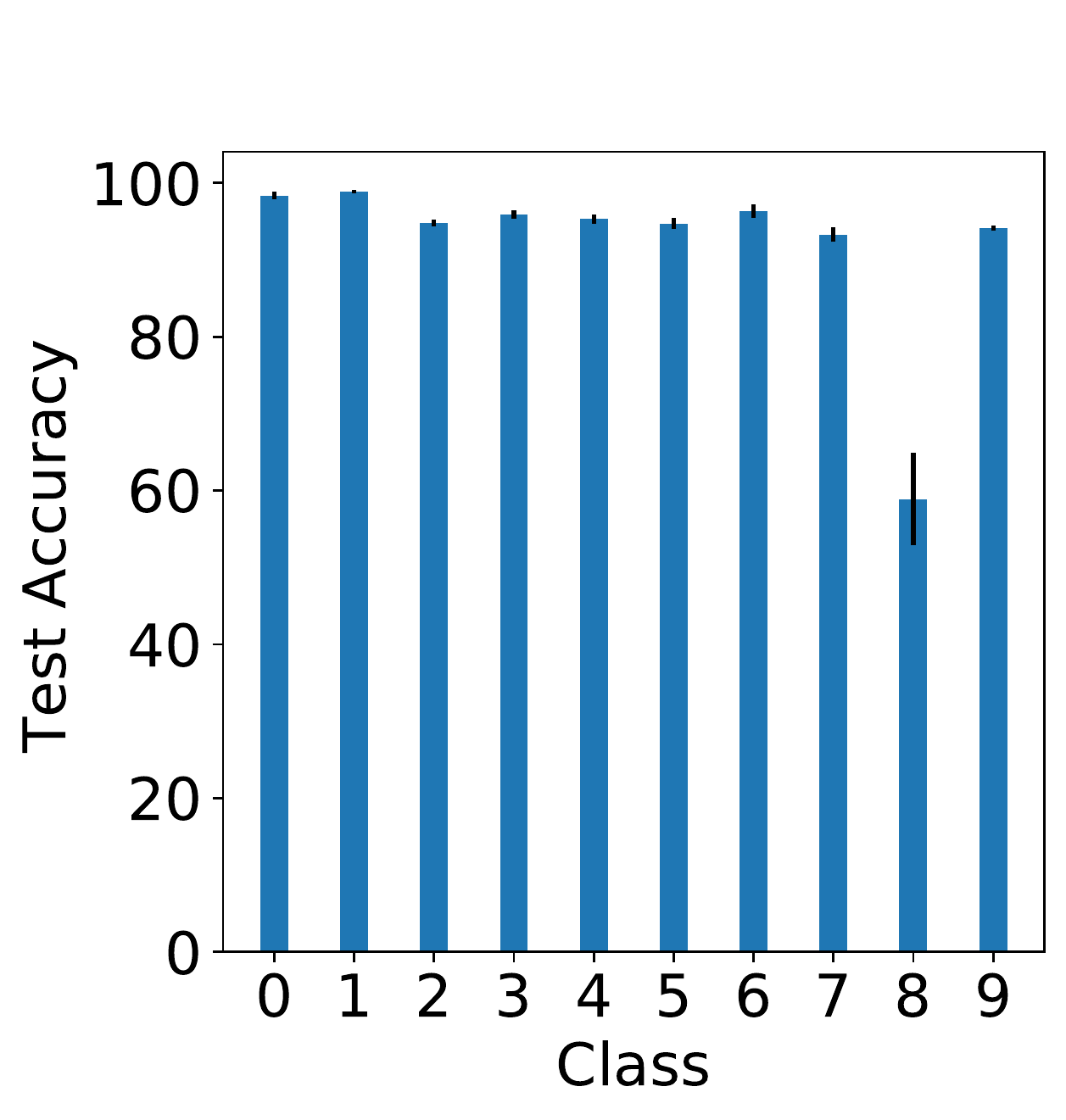}
     \caption{PATE for $\varepsilon=15$}
     \label{fig:mnistpateeps15}
    \end{subfigure}
    \caption{Average test accuracy of each digit (class) for models trained on imbalanced MNIST data, where samples of class ``8'' are decreased to 0.1 their original count. }
    \label{fig:mnistresults}
    \vspace{-2ex}
\end{figure*}

\begin{figure*}
    \centering
        \begin{subfigure}{0.22\textwidth}
     \includegraphics[width=\linewidth]{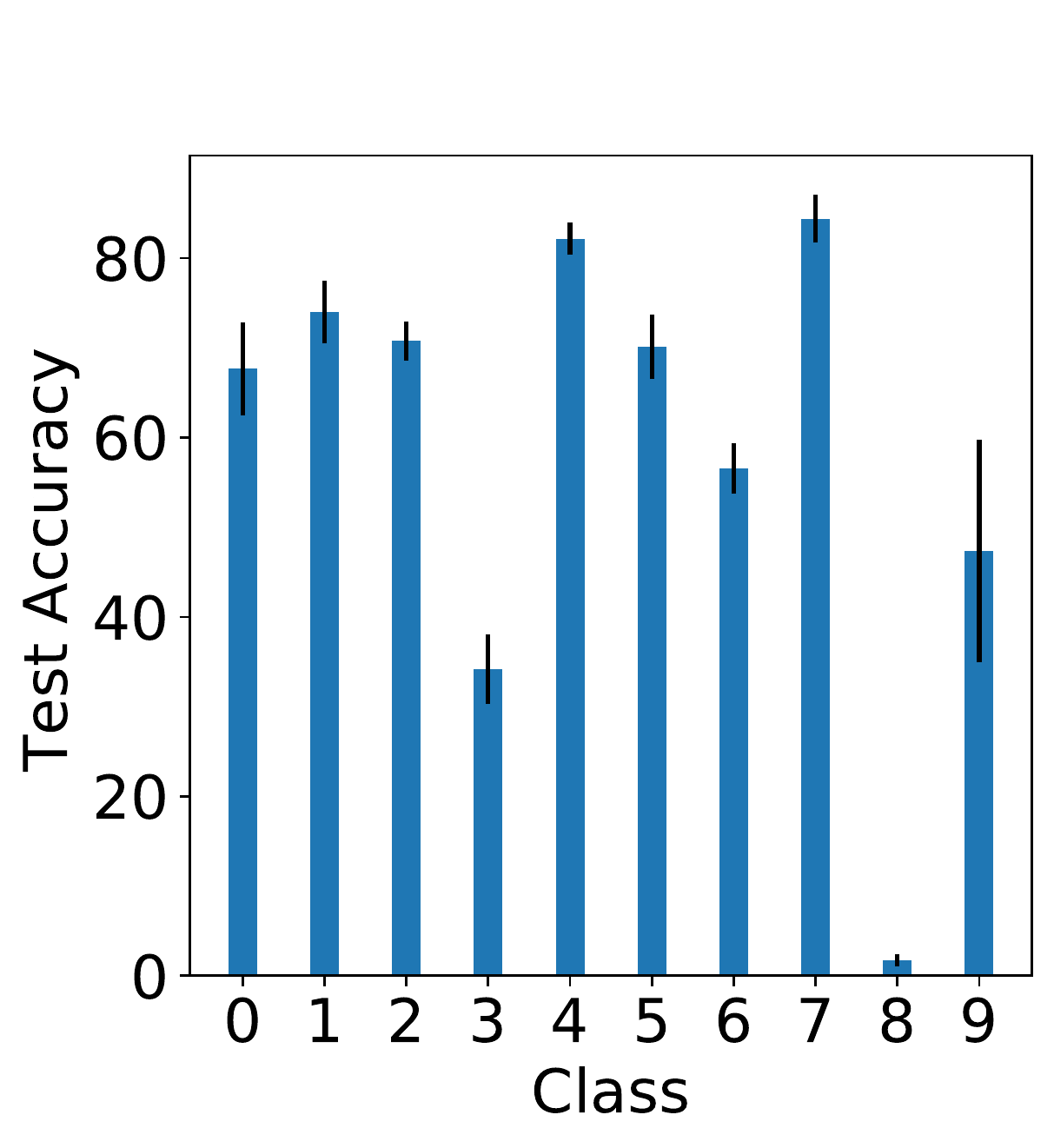}
     \caption{DP-SGD for $\varepsilon=5$}
     \label{fig:svhndpsgdeps5}
    \end{subfigure}
    \begin{subfigure}{0.22\textwidth}
     \includegraphics[width=\linewidth]{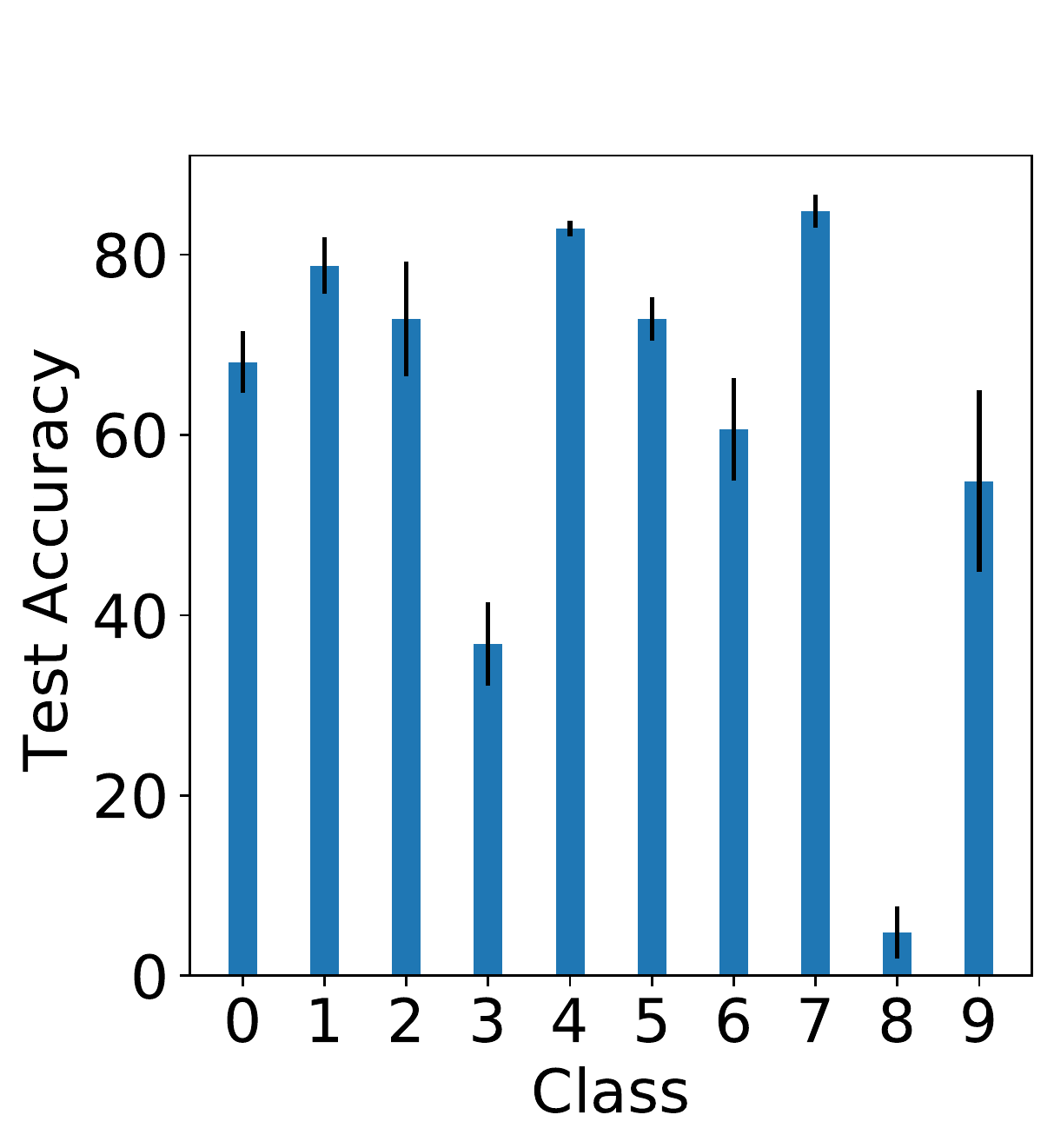}
     \caption{DP-SGD for $\varepsilon=8$}
     \label{fig:svhndpsgdeps8}
    \end{subfigure}
    \begin{subfigure}{0.22\textwidth}
     \includegraphics[width=\linewidth]{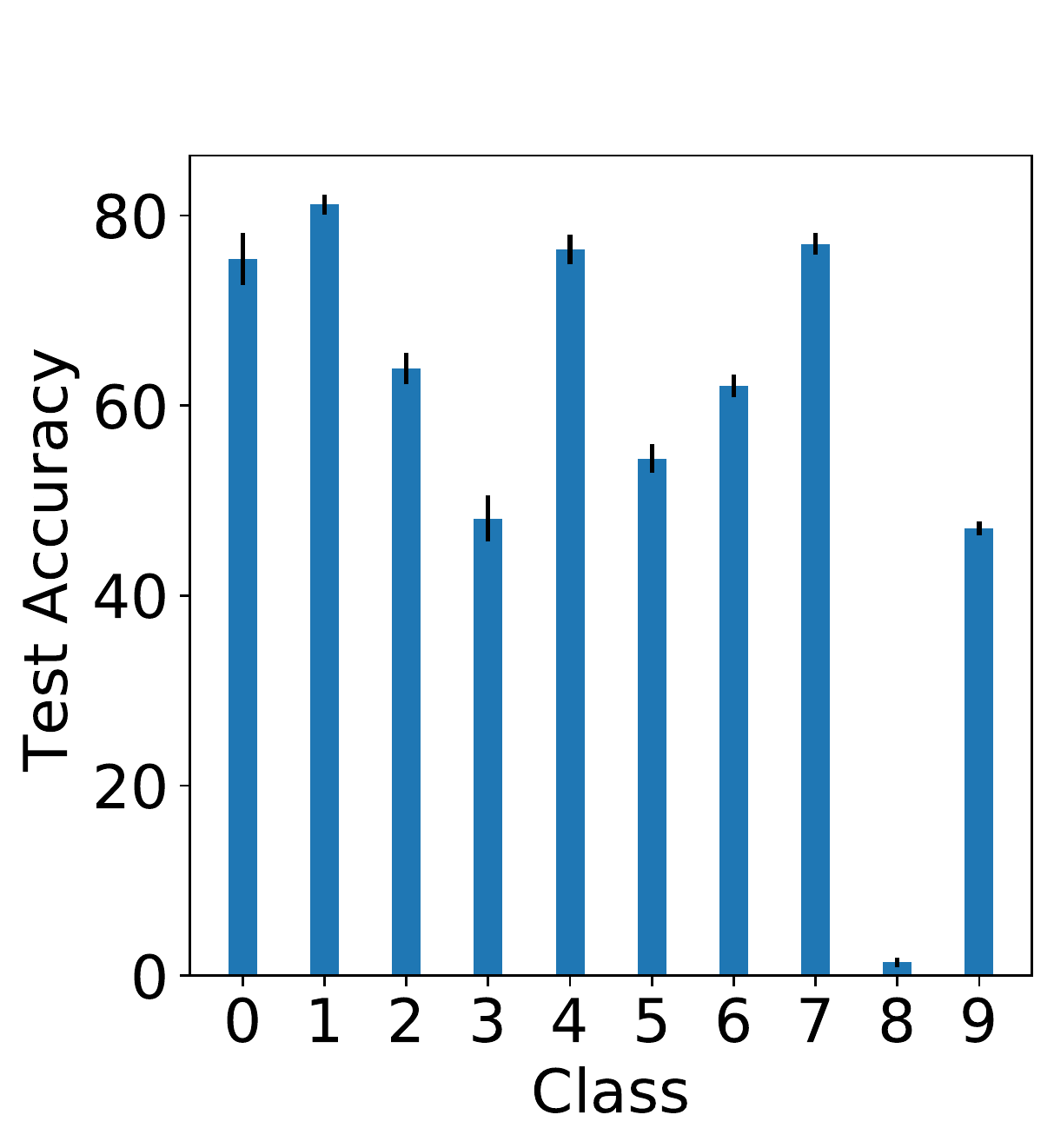}
     \caption{PATE for $\varepsilon=5$}
     \label{fig:svhnpateeps5}
    \end{subfigure}
    \begin{subfigure}{0.22\textwidth}
     \includegraphics[width=\linewidth]{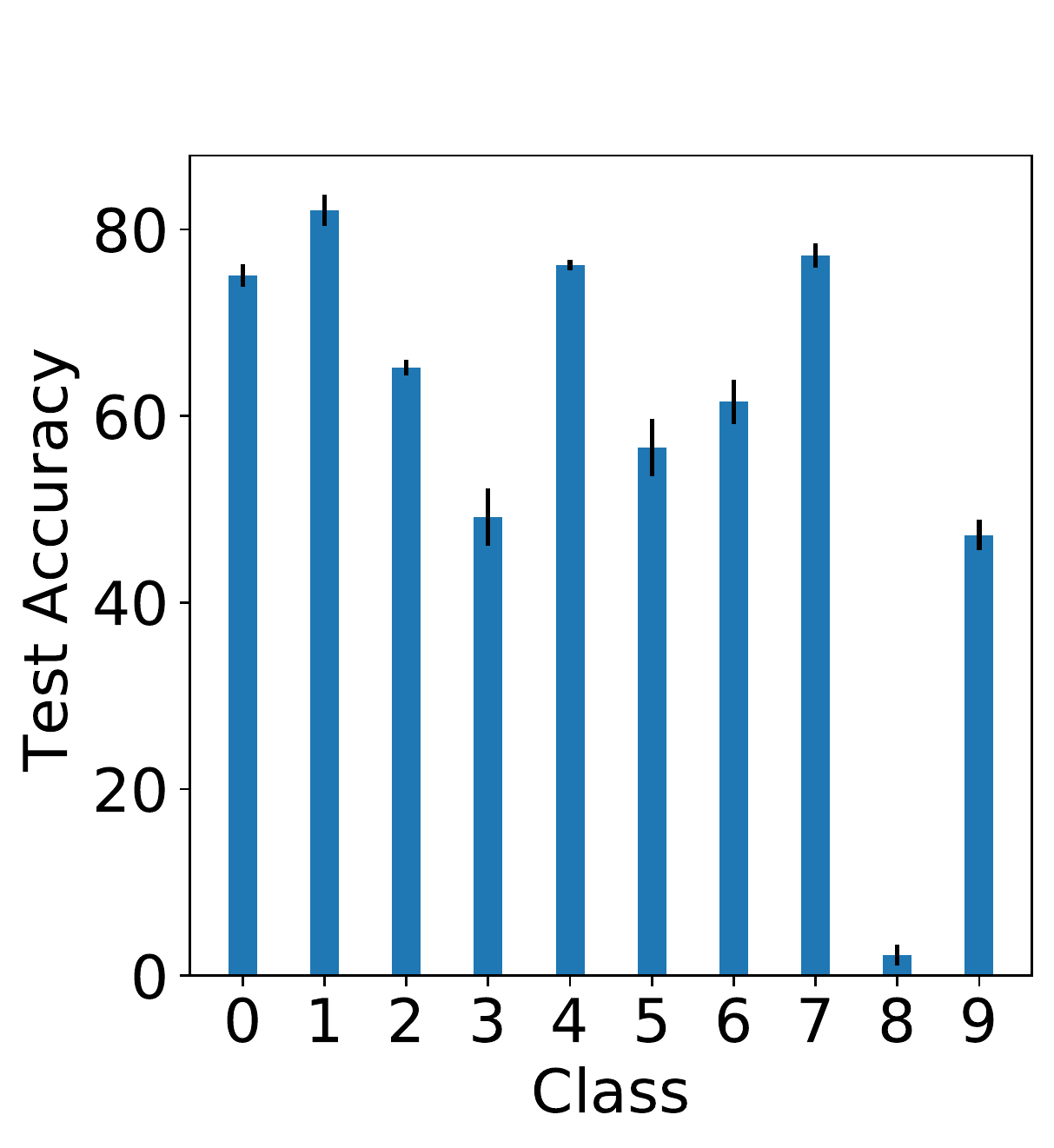}
     \caption{PATE for $\varepsilon=8$}
     \label{fig:svhnpateeps8}
    \end{subfigure}
    \caption{Average test accuracy of each digit (class) for models trained on imbalanced SVHN data, where samples of class ``8'' are decreased to half their original count. }
    \label{fig:svhnresults}
    \vspace{-2ex}
\end{figure*}

\begin{figure}[h]
    \centering
        \begin{subfigure}{0.4\textwidth}
     \includegraphics[width=\linewidth]{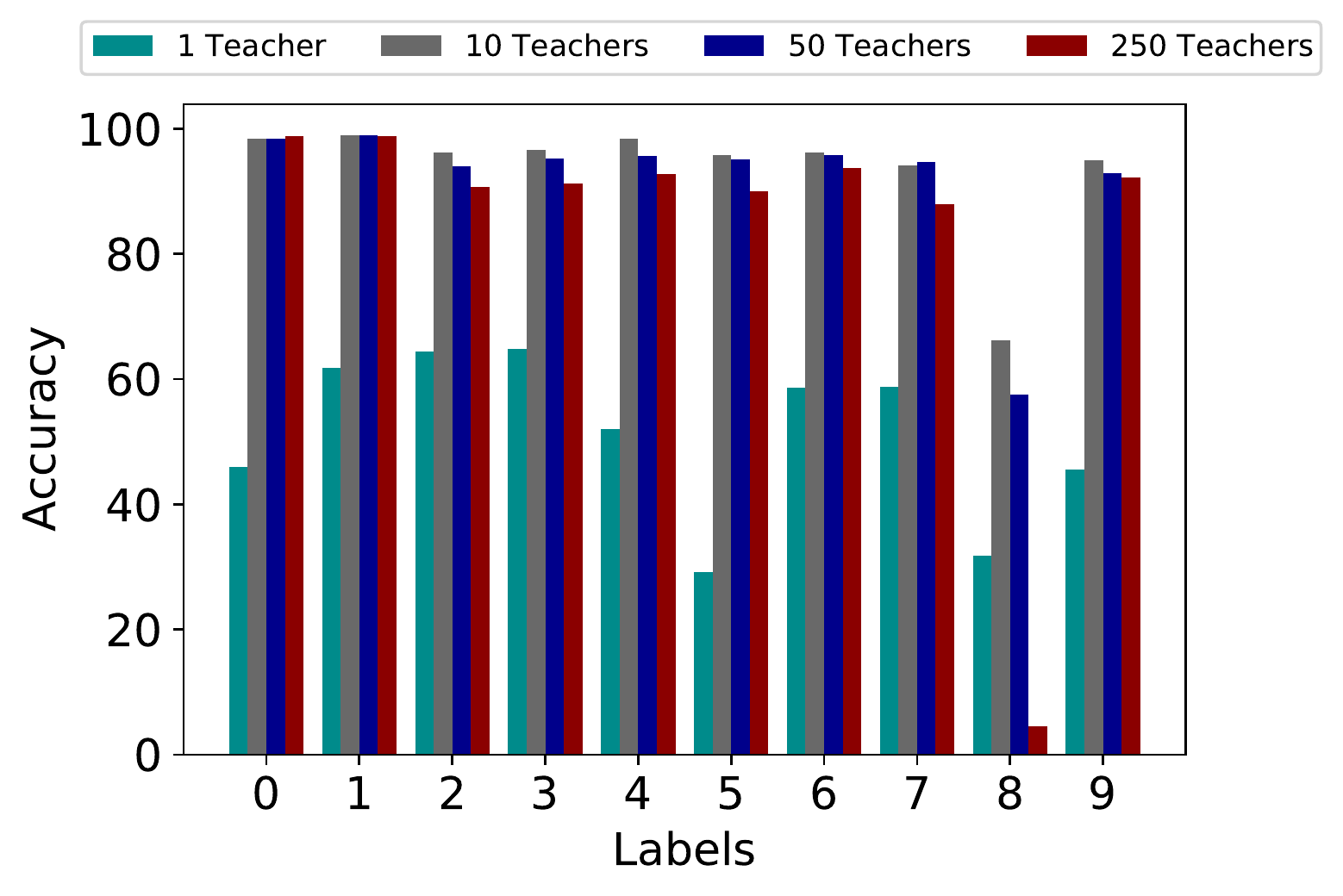}
     \caption{MNIST w/ PATE for $\varepsilon=5$}
     \label{fig:mnistablation}
    \end{subfigure}
    
    \begin{subfigure}{0.4\textwidth}
     \includegraphics[width=\linewidth]{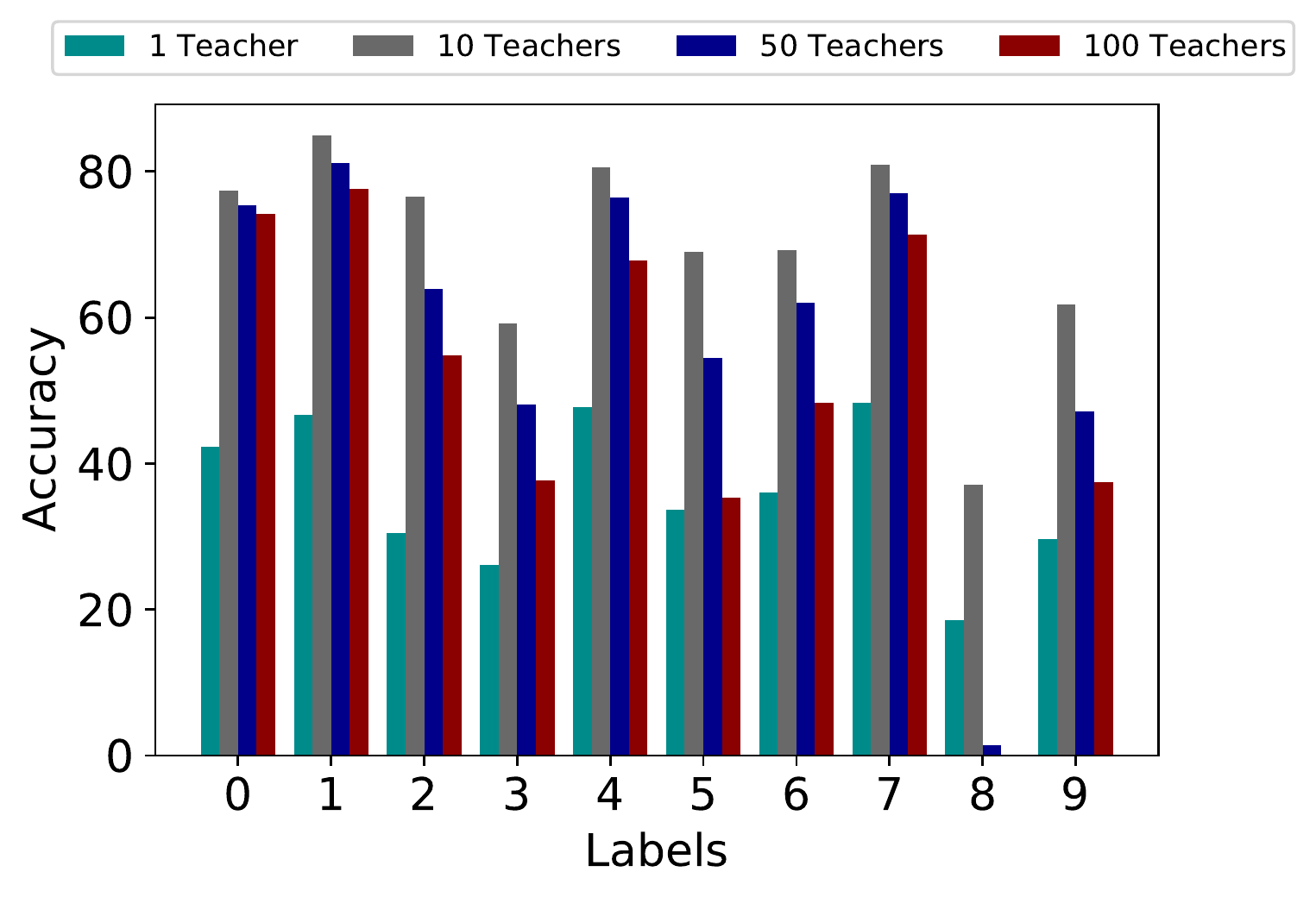}
     \caption{SVHN w/ PATE for $\varepsilon=5$}
     \label{fig:svhnablation}
    \end{subfigure}
    \caption{Effect the number of teachers has on accuracy of each class for  PATE.}
    \label{fig:ablationpate}
    \vspace{-2ex}
\end{figure}

\section{Empirical Evaluations}

To compare the effect that the two differentially private deep learning algorithms, DP-SGD and PATE have on sub-populations with different sizes, we run a series of experiments on MNIST and SVHN datasets. We have selected these two datasets, as they are the ones used for the evaluations of the PATE framework~\cite{papernot2016semi}.

\subsection{Experimental Setup}
For MNIST dataset, we artificially imbalance the class ``8'', to have only 500 samples. For SVHN we imbalance the same class as well. We decide to imbalance class $8$ to maintain consistency with \cite{bagdasaryan2019differential}. The ratio of imbalance is set to 1:10 i.e for every 10 images in each class, there was only 1 image in class 8. We apply this imbalance for the train set as well as the test set. In total, we sample 5000 training and 1000 test samples for each class except class 8. Class 8 gets 500 training and 100 test samples, respectively.
For SVHN, as the dataset~\cite{svhn11} is already  imbalanced (i.e. class $1$ and $2$ have nearly 4$\times$ training images than other classes causing the vote counts biased towards these classes in PATE), we balance it with each class containing $5000$ images \footnote{Classes with training images $< 5000$ were left untouched.}. 
We imbalance class $8$ by performing a $1:2$ imbalance i.e. for every $2$ images in each class, only $1$ image in class $8$. Therefore, class $8$ is alloted $2500$ training examples while other classes comprise of $5000$ images each. 

%

We use a 4-layer deep CNN for performing our experiments on MNIST dataset. The reason for using such a small CNN for MNIST is that the dataset contains grayscale images and performing feature extraction on such images is quite easier in comparison to RGB images.
%
%
For SVHN dataset, 
we utilized the ResNet-18 architecture \cite{resnet18}.
we use learning rates 0.01 and 0.05 for DP-SGD and PATE, respectively. We train PATE with 50 teachers and 30 students. We train each model 5 times and report the mean and standard deviation of accuracy in our experiments. 


\subsection{MNIST Results}
Figure~\ref{fig:mnistresults} displays test accuracies on the imbalanced MNIST dataset for $\epsilon=0.5,5,15$. We observe that the deviation of the test accuracy for the imbalanced class $8$ decreased with the increase in the values of $\epsilon$ in DP-SGD (refer Figures~\ref{fig:mnistdpsgdeps0.5}, \ref{fig:mnistdpsgdeps5}, \ref{fig:mnistdpsgdeps15}). We also notice that over different values of $\epsilon$, PATE exhibits ``stable'' results for the test accuracy of the imbalanced class. In PATE, the accuracy of the imbalanced class is almost (more than) twice the accuracy obtained by DP-SGD for the same $\epsilon$ value (see Figures~\ref{fig:mnistpateeps0.5}, \ref{fig:mnistpateeps5}, \ref{fig:mnistpateeps15}). 
%
 %
 We hypothesize that this is because (1) DP-SGD adds noise during every update of the model, destroying the gradient signals from underrepresented groups. This doesn't happen during training of the teachers. (2) teachers are not learning exactly the same patterns (due the data distribution). Therefore the set of teachers is diverse. This diversity allows to cancel biases among teaches, ie. aggregating decisions of different teachers may help to reduce the confusion to classify underrepresented class.~\citeauthor{grgic2017fairness} shows theoretically that diversity is key of achieving fairness.

\subsection{SVHN Results}
Figure~\ref{fig:svhnresults} showcases the test accuracies on the imbalanced SVHN dataset for $\epsilon=5,8$. We experiment with these two $\epsilon$ values as demonstrated in the \citet{papernot2016semi} paper. We observe that both DP-SGD and PATE produce similar results with the averaged accuracies for the imbalanced class being $< 5\%$. However, it is important to note that PATE gives rise to robust results while DP-SGD exhibits higher standard deviations over different values of $\epsilon$ (refer Fig.~\ref{fig:svhndpsgdeps5}, \ref{fig:svhndpsgdeps8}, \ref{fig:svhnpateeps5}, \ref{fig:svhnpateeps8}). This property can again be referred to how the ``students'' learn different patterns on the disjoint data present within ``teachers''.

\subsection{Ablation Study: Number of Teachers}
We analyze the impact of the number of teachers on the accuracy of different classes of PATE.
Figure~\ref{fig:ablationpate} shows test accuracy for different numbers of teachers ($n = 1, 10, 50, 100, 250$).
For $n = 1$, PATE shows a drop in the accuracy as the results are obtained from only one teacher, which makes them noisy and less robust, hampering the performance of the model.
For $n = 10$, PATE shows the best performance considering the size of the datasets we're using. The number of teachers seem to be optimal,  making it the most fair outcome.
For $n = 50$, the performance of PATE decreases overall and it could be observed that unfairness of the model increased for the imbalanced class 8.
For $n = 250$ and $n = 100$ for MNIST and SVHN respectively, we observed that the accuracy of the imbalanced class 8 was very low (close to 0\% accuracy) as the number of training set samples that could be distributed among the number of teachers became very small, causing the teachers not to train. 
The main take-away from this experiment is that there is a sweet spot for the number of teachers, and going up or going low would not necessarily have a positive or negative impact on accuracy of the sub-classes.

\section{Conclusion and Future Work}

Our work provides a comprehensive analysis of the comparison of utility provided across DP-SGD and PATE on imbalanced datasets. To summarize: 
\begin{enumerate}
    \item Even though both DP-SGD and PATE have disparate impact on the under-represented groups i.e.``Poor get poorer" - PATE has significantly less disproportionate impact on utility compared to DP-SGD. 
    \item We note that the standard deviation of the accuracy for each class over 5 runs was much lower in PATE compared to DP-SGD.
    \item By experimenting with various teacher counts, we observe that having multiple teachers often provides a higher utility than a single teacher for under-represented groups. However beyond the tipping point of this ensemble (10 teachers in our case), the utility stagnates and then starts dropping significantly. 
\end{enumerate}
%
%

It is worth noting, however, that although PATE has the above advantages, it is a semi-supervised approach, in which part of the training is done on publicly available, unlabeled data. Therefore, PATE assumes such data (even if very small in size) is available. In situations where we do not have access to such data, for instance for medical purposes where no patient note is released, PATE cannot be applied.

\newpage
\bibliography{main}
\bibliographystyle{icml2021}


\end{document}